\documentclass{article}

\usepackage{microtype}
\usepackage{graphicx}
\usepackage{subfigure}
\usepackage{booktabs} 

\usepackage{hyperref}

\usepackage{graphicx}
\usepackage{amsmath}
\usepackage{algorithm}
\usepackage{algorithmic}
\usepackage{amsfonts}       
\usepackage{nicefrac}       
\usepackage{microtype}      
\usepackage{xcolor}         
\usepackage{enumitem}
\usepackage{makecell}
\usepackage{bbding}
\usepackage{pifont}
\usepackage{amsthm}
\usepackage{multirow}
\usepackage{colortbl}

\usepackage[accepted]{icml2025}

\usepackage{amsmath}
\usepackage{amssymb}
\usepackage{mathtools}
\usepackage{amsthm}

\usepackage[capitalize,noabbrev]{cleveref}

\theoremstyle{plain}

\theoremstyle{definition}

\theoremstyle{remark}

\definecolor{gray}{RGB}{222,222,222}

\usepackage[textsize=tiny]{todonotes}

\icmltitlerunning{One-Shot Heterogeneous Federated Learning with Local Model-Guided Diffusion Models}

\begin{document}

\twocolumn[
\icmltitle{One-Shot Heterogeneous Federated Learning with\\Local Model-Guided Diffusion Models}




\begin{icmlauthorlist}
\icmlauthor{Mingzhao Yang}{fudan}
\icmlauthor{Shangchao Su}{fudan}
\icmlauthor{Bin Li \footnotemark[2]}{fudan}
\icmlauthor{Xiangyang Xue}{fudan}
\end{icmlauthorlist}

\icmlaffiliation{fudan}{Shanghai Key Laboratory of Intelligent Information Processing, School of Computer Science, Fudan University}

\icmlcorrespondingauthor{Bin Li}{libin@fudan.edu.cn}

\icmlkeywords{Machine Learning, ICML}

\vskip 0.3in
]



\printAffiliationsAndNotice{}  

\begin{abstract}

In recent years, One-shot Federated Learning (OSFL) methods based on Diffusion Models (DMs) have garnered increasing attention due to their remarkable performance. However, most of these methods require the deployment of foundation models on client devices, which significantly raises the computational requirements and reduces their adaptability to heterogeneous client models compared to traditional FL methods. In this paper, we propose FedLMG, a heterogeneous one-shot \textbf{Fed}erated learning method with \textbf{L}ocal \textbf{M}odel-\textbf{G}uided diffusion models. Briefly speaking, in FedLMG, clients do not need access to any foundation models but only train and upload their local models, which is consistent with traditional FL methods. On the clients, we employ classification loss and BN loss to capture the broad category features and detailed contextual features of the client distributions. On the server, based on the uploaded client models, we utilize backpropagation to guide the server’s DM in generating synthetic datasets that comply with the client distributions, which are then used to train the aggregated model. By using the locally trained client models as a medium to transfer client knowledge, our method significantly reduces the computational requirements on client devices and effectively adapts to scenarios with heterogeneous clients. Extensive quantitation and visualization experiments on three large-scale real-world datasets, along with theoretical analysis, demonstrate that the synthetic datasets generated by FedLMG exhibit comparable quality and diversity to the client datasets, which leads to an aggregated model that outperforms all compared methods and even the performance ceiling, further elucidating the significant potential of utilizing DMs in FL.
\end{abstract}

\section{Introduction}
\label{1}
Federated learning (FL)~\cite{mammen2021federated} has gained increasing attention recently. Standard FL relies on frequent communication between the server and clients. 
With the growing adoption of AI models by individual users, the application scenarios of FL have expanded significantly, including mobile photo album categorization and autonomous driving~\cite{nguyen2022deep,fantauzzo2022feddrive}. However, in these scenarios, the substantial communication cost associated with FL is often impractical for individual users. As a result, one-shot FL (OSFL) has emerged as a solution~\cite{DBLP:conf/ijcai/LiHS21, zhou2020distilled}. OSFL aims to establish the aggregated model within a single communication round. Currently, mainstream OSFL methods can be categorized into four types: {1) Methods using the auxiliary dataset~\cite{guha2019one,li2020practical,lin2020ensemble}}. {2) Methods training generators~\cite{zhang2022dense,heinbaugh2022data}}. {3) Methods transferring auxiliary information~\cite{zhou2020distilled,su2023one}}. {4) Methods based on DMs~\cite{yang2023exploring, zhang2023federated}}.

However, existing methods are hard to apply in real-world scenarios due to the following reasons: 1) Collecting public datasets on the server that comply with all client distribution is impractical, owing to privacy concerns and data diversity issues. 2) Due to the limited computational power and data of the clients, training generators on realistic client images is challenging. So most OSFL methods can only be applied to small-scale toy datasets, such as MNIST and CIFAR10. 3) The transmission of auxiliary information incurs communication costs, and extracting the auxiliary information on the clients also incurs additional computation costs, further restricting the practicality of OSFL. 4) Current OSFL methods that leverage DMs necessitate the deployment of foundation models on the clients~\cite{yang2023exploring, su2022cross}, such as CLIP~\cite{radford2021learning} and BLIP~\cite{li2023blip}, or directly involve client training of diffusion models~\cite{yang2024feddeo}, leading to significant communication and computational costs. In comparison to widely used traditional FL methods, which only require local training of client models, these additional burdens imposed by OSFL methods significantly increase the strain on client resources, thereby limiting their practical applicability. 

To reduce the computational requirements on client devices and further alleviate the burden on clients, we propose FedLMG, a heterogeneous one-shot \textbf{Fed}erated learning method with \textbf{L}ocal Model-\textbf{G}uided diffusion models. Our method involves using the locally trained client models to guide the DMs in generating the synthetic dataset that complies with different client distributions. Specifically, FedLMG consists of three steps: Local Training, Image Generation, and Model Aggregation. Firstly, based on the theoretical analyses of the client's local distribution and the server's conditional distribution, the clients independently train the client models on their private data and upload them to the server. Subsequently, assisted by the received client models, the server generates realistic images that comply with different client distributions based on DMs. After obtaining the synthetic dataset, we introduce three strategies to obtain the aggregated model: fine-tuning, multi-teacher distillation, and specific-teacher distillation. Through these strategies, we achieve the model aggregation in a single round of communication, without accessing any client data or any additional information transferring compared with standard FL.

To validate the performance of our method, we conduct extensive quantitation and visualization experiments on three large-scale real image datasets: DomainNet~\cite{peng2019moment}, OpenImage~\cite{kuznetsova2020open} and NICO++~\cite{zhang2022NICO++}. Sufficient quantitation experiments under various client scenarios demonstrate that our method outperforms all compared methods in a single communication round, and in some cases even surpasses the ceiling performance of centralized training, strongly underscoring the potential of DMs and providing convincing evidence for our aforementioned ideas. Visualization experiments also illustrate that our method generates synthetic datasets that comply with both the specific categories and the personalized client distributions, with comparable quality and diversity to the original client dataset. Moreover, we conduct thorough discussions on communication costs, computational costs, and privacy concerns, further enhancing the practicality of the proposed method.

In summary, this paper makes the following contributions:
\begin{itemize}
    \item We propose FedLMG, a novel OSFL method, to achieve real-world OSFL without utilizing any foundation models on the clients, ensuring no additional communicational or computational burden compared to traditional FL methods, thereby significantly expanding the practicality of OSFL.
    \item We propose using the locally trained client models as a medium to transfer client knowledge to the server, guiding the diffusion model through classification loss and BN loss to capture the client's category and contextual features, thereby generating high-quality synthetic datasets that comply with client distributions.
    \item We conduct thorough theoretical analyses, demonstrating that under the assistance of client models, the KL divergence between the data distribution of the DM on the server and the local data distribution is bounded. 
    \item We conduct sufficient quantitation and visualization experiments to demonstrate that the proposed method outperforms other compared methods and can even surpass the performance ceiling of centralized training in some cases, further evidencing the enormous potential of utilizing DM in OSFL.
\end{itemize}

\section{Related Work}

\subsection{One-shot Federated Learning}

In the standard FL~\cite{mcmahan2017communication}, there are multiple rounds of communication between the server and clients. To reduce the high communication costs, OSFL entails clients training their local models to convergence first, followed by aggregation on the server. Existing OSFL methods can be broadly categorized into three main types. {1) Methods based on public auxiliary dataset.} \cite{guha2019one} utilizes unlabeled public data on the server for model distillation. Similarly, FedKT~\cite{li2020practical} and FedDF~\cite{lin2020ensemble} employ an auxiliary dataset for knowledge transfer on the server. {2) Methods based on generators.} DENSE~\cite{zhang2022dense} employs an ensemble of client models as a discriminator to train a generator for generating pseudo samples, which is used to train the aggregated model. To address very high statistical heterogeneity, FedCVAE~\cite{heinbaugh2022data} trains a conditional variational autoencoder (CVAE) on the client side and sends the decoders to the server to generate data. {3) Methods based on sharing auxiliary information.} DOSFL~\cite{zhou2020distilled} performs data distillation on the client, and the distilled pseudo samples are uploaded to the server for global model training. MAEcho~\cite{su2023one} shares the orthogonal projection matrices of client features to the server to optimize global model parameters.


 \begin{figure*}[t]
 \centering
 \includegraphics[width=0.95\linewidth]{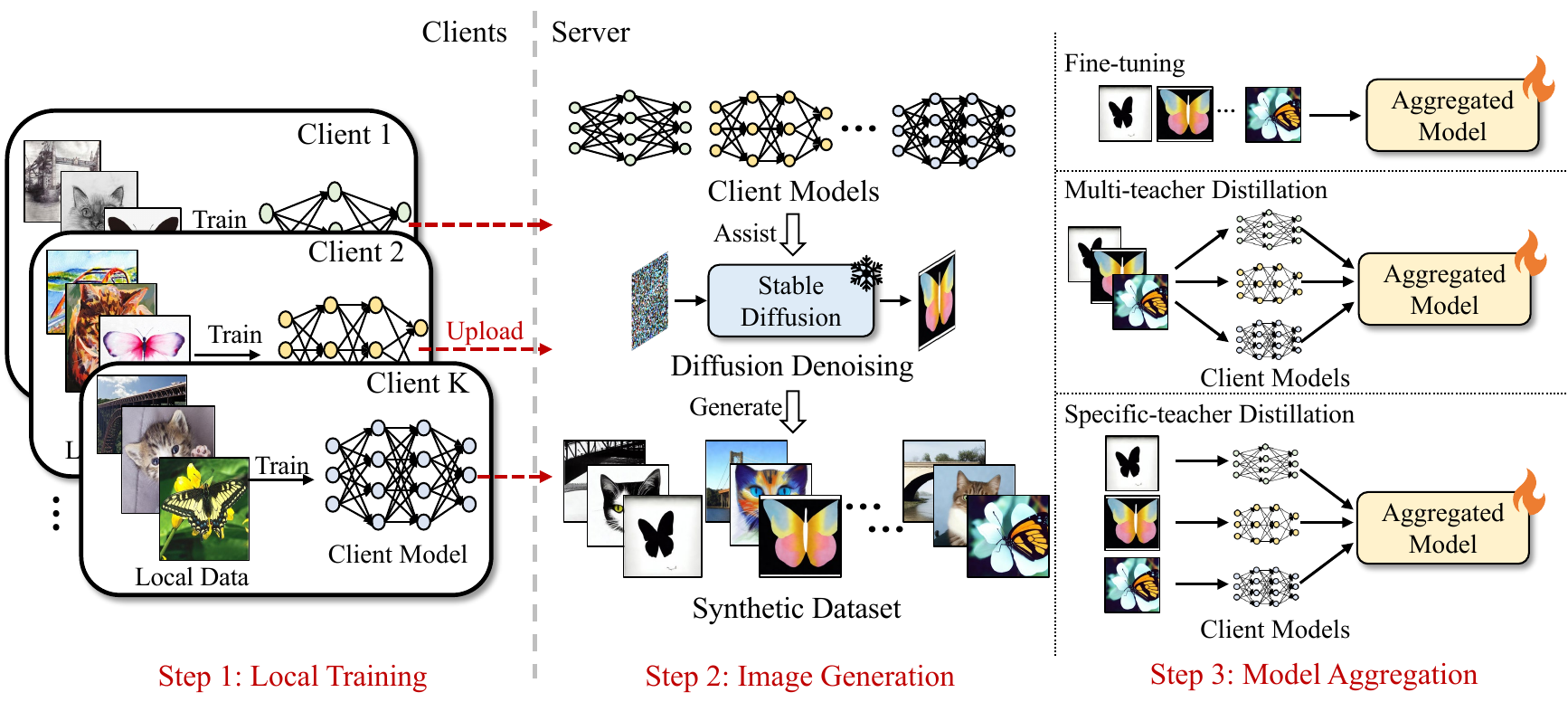}
 \caption{Overview of FedLMG. Our method consists of three steps: \textbf{Local Training}, \textbf{Image Generation}, and \textbf{Model Aggregation}. Firstly, each client independently trains their models using its private data and uploads them to the server. Assisted by these client models, our method leverages the powerful DM to obtain the synthetic dataset that complies with different client distributions. Based on the synthetic dataset, three strategies are provided to obtain the aggregated model.}
 \label{framework}
 \end{figure*}
 
  \subsection{FL with Pre-trained Diffusion Models}

Currently, only a small number of studies have focused on the significant potential of pre-trained DMs in FL. In FedDISC~\cite{yang2023exploring}, stable diffusion is introduced into semi-supervised FL for the first time, achieving remarkable results within just a single communication round. However, it requires using the CLIP image branch as the backbone for classification, which limits its flexibility. FGL~\cite{zhang2023federated} employ BLIPv2~\cite{li2023blip} on the client side to extract image descriptions, from which prompts are extracted and sent to the server for generation using DMs. This approach necessitates pre-deploying BLIPv2 on the client side. Phoenix~\cite{jothiraj2023phoenix} introduced FL into DMs, proposing a distributed method for training DMs. Unlike FedDISC and FGL, we do not require deploying any foundational models on the clients, which significantly reduces communication and computational costs on the clients, further enhancing the practicality of our method and enabling our method to address the scenario of heterogeneous client models.
 
\section{Method}
 
In this section, we introduce the proposed method in detail. Firstly, we provide essential preliminaries regarding the diffusion process and our problem setting. Then, as stated in Figure~\ref{framework}, we detail the three steps of the proposed method: Local Training, Image Generation, and Model Aggregation, including the theoretical analyses of the distributions of the synthetic data and the client's local data.

\subsection{Preliminaries}
\textbf{Problem Setting and Notation.}
Consider that we have $K$ clients. In OSFL, take client $k$ as an example, this client has a private dataset $\mathcal{D}^k=\left\{\left(\mathbf{x}_i,y_i\right)\right\}_{i=1}^{N_k}$ and a client model $\mathcal{F}_{\boldsymbol{\theta}_k}$. The server needs to aggregate the client models $\{\mathcal{F}_{\boldsymbol{\theta}_k}\}^K_{k=1}$ to obtain an aggregated model $\mathcal{F}_{\boldsymbol{\theta}_g}$ that adapts various client distributions. The overall objection of our framework is:
    \begin{equation}
        \min _{\mathbf{w} \in \mathbb{R}^{d}} \frac{1}{K} \sum_{k=1}^{K}  \mathbb{E}_{\mathbf{x} \sim \mathcal{D}_{k}}\left[\ell_{k}(\mathcal{F}_{\boldsymbol{\theta}_g} ; \mathbf{x})\right] 
    \end{equation}
where $\ell_{k}$ is the local objective function for the $k$-th client, $\mathcal{F}_{\boldsymbol{\theta}_g}$ is the parameters of the aggregated model. From this objective function, it is evident that our goal is to train an aggregated model that adapts to all client distributions and exhibits excellent classification performance on the data from each client. We also assess the model performance in subsequent experimental sections according to this objective.

\textbf{Diffusion Process.}
The DM $\epsilon_{\theta}$ samples initial noise ${\mathbf{s}}_T$ from a standard Gaussian Distribution $\mathcal{N}(0,\mathcal{I})$ and iteratively denoises it, resulting in a realistic image ${\mathbf{s}}_0$, where $T$ denotes the maximum timestep. \cite{dhariwal2021diffusion} proposes the classifier-guidance, wherein the gradients backpropagated through the classifier $\mathcal{F}_{\boldsymbol{\theta}_k}$ are used to modify the predicted noise of the DM $\epsilon_{\theta}(\mathbf{s}_{t},t)$ at each timestep. The loss function utilized for generating gradients typically involves the cross-entropy loss $\mathcal{L}_{CE}$ between the given class label $y$ and the output of classifier $\mathcal{F}_{\boldsymbol{\theta}_k}(\mathbf{s}_{t})$. This enables the DM to generate images with specified classes. With its textual prompt introduced as additional conditions, the sampling process at each time step has two steps. Firstly, for any given timestep $t\in\left \{ 0,\dots,T \right \} $, the predicted noise is modified according to the following equation:

\begin{equation}
\label{ftnoise}
\hat{\epsilon}\left(\mathbf{s}_{t},t|y\right):=\epsilon_{\theta}\left(\mathbf{s}_{t},t|y\right)-\sqrt{1-\bar{\alpha}_{t}} \nabla_{\mathbf{s}_{t}} \log p_{\boldsymbol{\theta}_k}\left(y|\mathbf{s}_{t}\right)
\end{equation}
Afterwards, utilizing the modified $\hat{\epsilon}\left(\mathbf{s}_{t},t|y\right)$, the sample for the next time step $\mathbf{s}_{t-1}$ is obtained:
    \begin{align}
    \label{diffusion}
     \mathbf{s}_{t-1} = &\sqrt{\alpha_{t-1}}\Big(\frac{\mathbf{s}_{t}-\sqrt{1-\alpha_{t}}\hat{\epsilon}(\mathbf{s}_{t},t|y)}{\sqrt{\alpha_{t}}}\Big) \nonumber \\
    &+\sqrt{1-\alpha_{t-1}-\sigma_{t}^{2}}\cdot\hat{\epsilon}(\mathbf{s}_{t},t|y)+\sigma_{t}\varepsilon _{t}
    \end{align}
 where $\alpha_{t}$, $\alpha_{t-1}$ and $\sigma_{t}$ are pre-defined parameters, $\varepsilon_{t}$ is the Gaussian noise randomly sampled at each timestep. 
According to the diffusion process described above, for the randomly sampled initial noise $\mathbf{s}_{T} \sim \mathcal{N}(0,\mathcal{I})$, after $T$ iterations, the diffusion model can denoise the initial noise into high-quality realistic images $\mathbf{s}_{0}$.

 \subsection{Local Training}
 \label{3.2}
 The first step of our method is local training. To generate data on the server that complies with the client distribution, we need to transmit client information to the server. In federated learning, it's common to send locally trained client models to the server. Therefore, for the local data distribution $p_{k}(\mathbf{x})$ of the private dataset $\mathcal{D}^k$ for client $k$, we leverage the information from the locally trained client model $\mathcal{F}_{\boldsymbol{\theta}_k}$ on the server and obtain the conditional distribution $p_{\epsilon_{\theta}}(\mathbf{x}|\boldsymbol{\theta}_k)$ based on the data distribution of the diffusion model $p_{\epsilon_{\theta}}(\mathbf{x})$. We aim for the synthetic dataset sampled from the conditional distribution $p_{\epsilon_{\theta}}(\mathbf{x}|\boldsymbol{\theta}_k)$ to closely comply with the client local data distribution $p_{k}(\mathbf{x})$. Therefore, considering the relationship between $p_{k}(\mathbf{x})$ and $p_{\epsilon_{\theta}}(\mathbf{x}|\boldsymbol{\theta}_k)$ is essential. We conduct comprehensive theoretical analyses regarding the relationship between these two distributions.
 
Firstly, we need to analyze the relationship between the unconditional data distribution of diffusion model $p_{\epsilon_{\theta}}(\mathbf{x})$ and the client local data distribution $p_{k}(\mathbf{x})$. As stated in the Introduction, our motivation for utilizing the diffusion model lies in its ability to generate data that comply with almost any data distribution with proper guidance. Therefore, regarding the data distribution $p_{k}(\mathbf{x})$ of the client's local dataset $\mathcal{D}^{k}$ and the data distribution $p_{\epsilon_{\theta}}(\mathbf{x})$ that the DMs $\epsilon_{\theta}$ can generate, we can make the following assumption:

\textbf{Assumption 1} \textit{There exists $\lambda > 0$ such that the Kullback-Leibler divergence from $p_{k}(\mathbf{x})$ to $p_{\epsilon_{\theta}}(\mathbf{x})$ is bounded above by $\lambda$:} 
\begin{equation}
\label{assumption 1}
KL(p_{\epsilon_{\theta}}(\mathbf{x})\|p_{k}(\mathbf{x}))< \lambda 
\end{equation}
This assumption is adaptable. We don't rigidly demand that the distribution of the diffusion model $p_{\epsilon_{\theta}}(\mathbf{x})$ fully encompasses the client local distributions $p_{k}(\mathbf{x})$. Instead, we merely assume some overlap between these distributions. Even if the clients specialize in certain professional domains, like medical images, it's entirely viable to train specialized diffusion models on the server. Hence, this assumption is entirely reasonable, considering a comprehensive assessment of practical scenarios. Based on Assumption 1, we have the following theorem regarding the relationship between the conditional distribution $p_{\epsilon_{\theta}}(\mathbf{x}|\boldsymbol{\theta}_k)$ and the client local data distribution $p_{k}(\mathbf{x})$:

\textbf{Theorem 1} \textit{There exists $\lambda > 0$, for the local data distribution $p_{k}(\mathbf{x})$ and the conditional distribution $p_{\epsilon_{\theta}}(\mathbf{x}|\boldsymbol{\theta}_k)$ of the DM $\epsilon_{\theta}$ conditioned the client model $\mathcal{F}_{\boldsymbol{\theta}_k}$ trained on client $k$, we have:}

\begin{align}
\label{theorem}
KL(p_{k}(\mathbf{x})\|p_{\epsilon_{\theta}}(\mathbf{x}|\boldsymbol{\theta}_k)) &<\lambda + \mathbb{E}(\log p_{\epsilon_{\theta}}(\boldsymbol{\theta}_k)) \nonumber \\
& -\int p_{k}(\mathbf{x}) \log p_{\epsilon_{\theta}}(\boldsymbol{\theta}_k|\mathbf{x})d\mathbf{x}
\end{align}
For a detailed proof, please refer to the appendix. From Eq.~\ref{theorem}, we can observe that the KL divergence between the conditional distribution $p_{\epsilon_{\theta}}(\mathbf{x}|\boldsymbol{\theta}_k)$, which is also the distribution of the synthetic data, and the distribution of client's local data $p_{k}(\mathbf{x})$ is bounded above. This upper bound consists of three components: $\lambda$, $\mathbb{E}(\log p_{\epsilon_{\theta}}(\boldsymbol{\theta}_k))$, and $-\int p_{k}(\mathbf{x}) \log p_{\epsilon_{\theta}}(\boldsymbol{\theta}_k|\mathbf{x})d\mathbf{x}$. $\lambda$ is the same as in Eq.~\ref{assumption 1}. $\mathbb{E}(\log p_{\epsilon_{\theta}}(\boldsymbol{\theta}_k))$ is a constant independent of the sample $\mathbf{x}$. $-\int p_{n}(\mathbf{x}) \log p_{\epsilon_{\theta}}(\boldsymbol{\theta}_k|\mathbf{x})d\mathbf{x}$ is the negative log-likelihood between the client model $\mathcal{F}_{\boldsymbol{\theta}_k}$ and the client distribution $p_{k}(\mathbf{x})$. Minimizing this negative log-likelihood is equivalent to minimizing the cross-entropy loss $\mathcal{L}_{CE}$. This implies that during the local training, we need to train client models using the cross-entropy loss to minimize this upper bound of the KL divergence. Consequently, the conditional distribution of the synthetic dataset $p_{\epsilon_{\theta}}(\mathbf{x}|\boldsymbol{\theta}_k)$ can closely approximate the client's local data distribution $p_{k}(\mathbf{x})$. Therefore, for client 
$k$, we utilize its privacy dataset $\mathcal{D}^k$ and train the client model $\mathcal{F}_{\boldsymbol{\theta}_k}$ using the following cross-entropy loss function:
\begin{equation}
    \ell_{k}(\mathcal{F}_{\boldsymbol{\theta}_g} ; \mathbf{x}) = \mathcal{L}_{CE}(\mathcal{F}_{\boldsymbol{\theta}_k}(\mathbf{x}^k_i),y^k_i)
\end{equation}
After multiple rounds of training, the client models are sent to the server to guide the generation process on the server. It's worth noting that in FedLMG, there are no requirements for the used model structures, which further enhances the practicality of our method.
 
\subsection{Image Generation}
\label{3.3}
After receiving the locally trained client models $\mathcal{F}_{\boldsymbol{\theta}_k}, k = 1, \dots, K$  uploaded by the clients, these client models serve as cues for the DM, generating the synthetic dataset complies with different client distributions. Firstly, we elaborate on how the client models assist DM in generation. In our problem setting, generated images must possess accurate categories and comply with specified client distributions, introducing novel demands to sampling, and necessitating consideration of additional image attributes such as style,  color, background, etc. Relying solely on classification results falls significantly short of achieving these demands since the classification results mainly provide information on categories.  
    
    To provide detailed context information about the client distributions, we utilize the statistics of each batch normalization (BN) layer of client models $\mathcal{F}_{\boldsymbol{\theta}_k}$: mean $\mu$ and variance $\sigma$. In other words, we need to consider the conditional generation process $p(\mathbf{s}_{t-1}|\mathbf{s}_t,y,\{\boldsymbol{\mu}_{k,l}\}^{L_k}_{l=1},\{\boldsymbol{\sigma}_{k,l}\}^{L_k}_{l=1})$, where $\boldsymbol{\mu}_{k,l}$ and $\boldsymbol{\sigma}_{k,l}$ respectively denote the means and the variances of all BN layers within $\mathcal{F}_{\boldsymbol{\theta}_k}$, and $L_k$ represents the number of BN layers within $\mathcal{F}_{\boldsymbol{\theta}_k}$. Therefore, during modifying the predicted noise of the DM $\epsilon_{\theta}(\mathbf{s}_{t},t|y)$ at each time step $t$, we compute gradients by summing the cross-entropy loss $\mathcal{L}_{CE}$ and the BN Loss $\mathcal{L}_{BN}$ to incorporate the additional distribution details embedded within the statistics of the BN layers into the diffusion process. The computation of $\mathcal{L}_{BN}$ is as follows:
    \begin{align}
    \label{BN loss}
    \sum_{l=1}^{L}(\left \| \boldsymbol{\mu}_l(s,\boldsymbol{\theta}_k) - \boldsymbol{\mu}_{k,l} \right \| + \left \| \boldsymbol{\sigma}^2_l(\mathbf{s},\boldsymbol{\theta}_k) - \boldsymbol{\sigma}^2_{k,l}  \right \| ) 
    \end{align}    
    where $\boldsymbol{\mu}_l(s,\boldsymbol{\theta}_k)$ and $\boldsymbol{\sigma}^2_l(\mathbf{s},\boldsymbol{\theta}_k)$ denote the mean and variance of the output feature from the $l$-th BN layer after feeding the sample $\mathbf{s}$ into the client model $\mathcal{F}_{\boldsymbol{\theta}_k}$.
    
    Furthermore, since client models are simply trained on the client $\mathcal{F}_{\boldsymbol{\theta}_k}$ and are not accustomed to the noised input $\mathbf{s}_{t}$, traditional classifier-guidance struggles to provide accurate guidance through the computed gradient $\nabla_{\mathbf{s}_{t}} \log p_{\boldsymbol{\theta}_k}\left(y|\mathbf{s}_{t}\right)$. To address this challenge, at any time step $t$, we utilize the predicted noise $\epsilon_{\theta}(\mathbf{s}_{t},t|y)$ to predict $\mathbf{s}_{0}$ according to the following equation:
    \begin{equation}
    \hat{\mathbf{s}}_{0,t} =\frac{\mathbf{s}_{t}-\sqrt{1-\alpha_{t}}\hat{\epsilon}(\mathbf{s}_{t},t|y)}{\sqrt{\alpha_{t}}}
    \end{equation}    
   Subsequently, based on $\hat{\mathbf{s}}_{0,t}$, we compute the loss function and gradient to modify $\epsilon_{\theta}(\mathbf{s}_{t},t|y)$. Although in the initial time steps, $\hat{\mathbf{s}}_{0,t}$ may appear blurry, the noise level in comparison to $\mathbf{s}_{t}$ is noticeably reduced. This decreases the demand for the client models' robustness of noise, mitigating the need for clients to specifically train for classifying noised samples and further enhancing the practicality of our method.
   
   Finally, the overall loss function $\mathcal{L}$ employed in the conditional generation is as follows:
   \begin{equation}
    \label{ce+bn loss}
        \mathcal{L}(\mathbf{s}_{t},y,\theta_k) = \mathcal{L}_{CE}(\mathcal{F}_{\boldsymbol{\theta}_k}(\hat{\mathbf{s}}_{0,t}),y)+\lambda \mathcal{L}_{BN}(\hat{\mathbf{s}}_{0,t},\theta_k )
   \end{equation}
where $\lambda$ is the weight of BN Loss. After performing gradient backpropagation according to this loss function, we use Eq.~\ref{ftnoise} and~\ref{diffusion} to guide the generation process through the client model $\mathcal{F}_{\boldsymbol{\theta}_k}$ and its accompanying BN statistics, enabling conditional generation. 

 For any given time step $t\in\left \{ 0,\dots, T\right \}$, we modify the predicted noise of the DM based on Eq.~\ref{ce+bn loss}:
\begin{equation}
\hat{\epsilon}\left(\mathbf{s}_{t},t|y\right):=\epsilon_{\theta}\left(\mathbf{s}_{t},t|y\right)-\sqrt{1-\bar{\alpha}_{t}} \nabla_{\mathbf{s}_{t}} \mathcal{L}(\mathbf{s}_t,y,\theta_k)
\end{equation}
Subsequently, based on Eq.~\ref{diffusion}, we compute $\mathbf{s}_{t-1}$ using the modified $\hat{\epsilon}\left(\mathbf{s}_{t},t|y\right)$, leading to the realistic image $\mathbf{s}_{0}$ after $T$ iterations. During the generation, since we specify the category $y$ and the classifier $\mathcal{F}_{\boldsymbol{\theta}_k}$, the generated image $\mathbf{s}_{0}$ is automatically labeled. We define $\mathbf{s}_{0}$ as $\hat{\mathbf{x}}^k_i$ and include along with its label $y^k_i$ in the synthetic dataset $\hat{\mathbf{X}}$. After undergoing multiple iterations of generation, we obtain the synthetic dataset $\hat{\mathbf{X}}= \{{(\hat{\mathbf{x}}^k_i},y^k_i)\}^N_{i=1}$.
 \begin{table*}[t]
 \caption{Performance of different methods on OpenImage, DomainNet, Unique NICO++, and Common NICO++ under feature distribution skew, where italicized text represents the ceiling performance used solely as a reference, and bold text signifies the optimal performance excluding the ceiling performance.}
\label{fea_skew}
\center
\resizebox{0.9\linewidth}{!}{
\begin{tabular}{ccccccccccccccc}
\Xhline{1.2pt}
                                               & \multicolumn{7}{c}{OpenImage}                                                                                     & \multicolumn{7}{c}{DomainNet}                                                                                           \\ 
                                               & client0        & client1        & client2        & client3        & client4        & client5        & \multicolumn{1}{c|}{Avg   }         & clipart        & infograph      & painting    &quickdraw   & real           & sketch        & \multicolumn{1}{c}{Avg}            \\ \hline
\textit{Ceiling}                & \textit{49.88}          & \textit{50.56}          & \textit{57.89}          & \textit{59.96}          & \textit{66.53}          & \textit{51.38}          &\multicolumn{1}{c|}{ \textit{56.03}    }      & \textit{47.48}          & \textit{19.64}          & \textit{45.24}          & \textit{12.31}          & \textit{59.79}  & \textit{42.35}        & \multicolumn{1}{c}{\textit{36.89}}          \\ \hline 
FedAvg        & 41.46          & 50.36          & 52.61          & 50.36          & 62.10          & 50.17          &\multicolumn{1}{c|}{  51.18}          & 37.96 & 12.55 & 34.41 & 5.93 & 51.33 & 32.37 & 29.09  \\
FedDF         & 44.96          & 46.15          & 59.69          & 58.69          & 63.45          & 46.63          & \multicolumn{1}{c|}{ 53.26}          & 38.09 & 13.68 & 35.48 & 7.32 & 53.83 & 34.69 & 30.52   \\
 FedProx         & 44.99        & 48.83 & 49.25 & 56.68 & 61.23 & 46.07  & \multicolumn{1}{c|}{ 51.18}          & 38.24 & 12.46 & 37.29 & 6.26 & 54.88 & 35.76 & 30.82      \\ 
 FedDyn         & 46.93          & 46.08 & 52.44 & 54.67 & 62.84 & 47.73 & \multicolumn{1}{c|}{ 51.78}          &  40.12 & 14.77 & 36.59 & 7.73 & 54.85 & 34.81 & 31.48         \\ \hline
Prompts Only        & 30.41          & 30.23 & 42.92 & 43.48 & 50.75 & 33.43 & \multicolumn{1}{c|}{ 38.54}          & 31.8  & 11.61 & 31.14 & 4.13 & \textbf{61.53} & 31.44 & 28.61  \\
FedDISC & 47.42 & 49.65 & 54.73 & 53.41 & 60.74 & 52.81 & \multicolumn{1}{c|}{ 53.13}          & 43.89 & 14.84 & 38.38 & 8.35 & 56.19 & 36.82 & 33.08  \\
FGL     & 48.21 & 49.16 & 54.98 & 55.47 & 63.14 & 49.32 & \multicolumn{1}{c|}{ 53.38}          &  41.81 & 15.30  & 40.67 & 8.79 & 57.58 & 39.54 & 33.95         \\ \hline
 FedLMG\_FT & \textbf{48.99} & 51.66          & 55.59          & 52.80          & 62.41          & 58.86          & \multicolumn{1}{c|}{ 55.05  }        & 44.25 & 17.51 & 38.74 & 9.43  & 57.31 & 38.44 & \multicolumn{1}{c}{34.28}          \\
FedLMG\_SD     & 47.60          & \textbf{55.20} & \textbf{61.54} & \textbf{61.83} & \textbf{67.07} & \textbf{59.90} & \multicolumn{1}{c|}{ \textbf{58.86}} & \textbf{46.23} & 18.42 & \textbf{42.85} & \textbf{10.24} & 58.52 & 39.13 & 35.90          \\
FedLMG\_MD     & 44.70          & 53.08          & 58.67          & 60.13          & 64.06          & 58.06          & \multicolumn{1}{c|}{ 56.45 }         & 47.21    & \textbf{18.49} & 40.37 & 10.02 & 59.67 & \textbf{40.19} & \textbf{35.99}  \\ \Xhline{1.2pt}
                                    &               & \multicolumn{6}{c}{Unique NICO++}                                                                                    & \multicolumn{7}{c}{Common NICO++}                                                                                       \\               & client0        & client1        & client2        & client3        & client4        & client5        & \multicolumn{1}{c|}{ Avg }           & autumn         & dim            & grass          & outdoor        & rock           & water            & Avg             \\ \hline
\textit{Ceiling}               & \textit{79.16}          & \textit{81.51}          & \textit{76.04}          & \textit{72.91}          & \textit{79.16}          & \textit{79.29}          & \multicolumn{1}{c|}{ \textit{78.01} }         & \textit{62.66}          & \textit{54.07}          & \textit{64.89}          & \textit{63.04}          & \textit{61.08}          & \textit{54.63}            & \textit{60.06}           \\ \hline
FedAvg        & 67.31          & 74.73          & 69.01          & 64.37          & 73.07          & 67.87          & \multicolumn{1}{c|}{ 69.39  }        & 52.51          & 40.45          & 57.21          & 51.59          & 49.31          & 43.56            & 49.11           \\
FedDF         & 69.79          & 78.90          & 69.53          & 66.01          & 74.86          & 70.80          & \multicolumn{1}{c|}{ 71.65  }        & 50.44          & 39.62          & 57.42          & 52.91          & 51.61          & 44.76            & 49.46           \\ 
FedProx         & 70.46          & 75.3         & 70.87          & 67.67          & 72.84         & 71.51          & \multicolumn{1}{c|}{ 71.44  }        & 53.49          & 42.41          & 58.84         & 53.08          & 53.67          & 45.42            & 51.15           \\
FedDyn         & 71.23          & 74.98          & 69.68          & 68.13          & 73.63          & 70.61          & \multicolumn{1}{c|}{ 71.37  }        & 54.38          & 43.20          & 57.56          & 52.63          & 52.86          & 46.76            & 51.23           \\\hline
Prompts Only       & 69.79          & 69.14          & 69.32          & 59.89          & \multicolumn{1}{c}{ 67.70}          & 66.60          & \multicolumn{1}{c|}{ 67.07   }       & 50.51          & 38.10          & 54.53          & 49.39          & 49.12          & 41.58            & 47.21           \\
FedDISC & 74.32 & 73.47 & 71.25 & 66.79 & 75.28 & 70.06 & \multicolumn{1}{c|}{ 71.86}          & 56.82 & 51.43 & 59.45 & 56.17 & 52.32 & 45.64 & 53.64 \\
FGL     & 74.62 & 79.43 & 71.26 & 68.65 & 76.37 & 74.31 & \multicolumn{1}{c|}{ 74.11}          &  57.25 & 49.35 & 61.81 & 58.42 & 54.29 & 47.62 & 54.79        \\ \hline
FedLMG\_FT & 75.13          & 73.30          & 70.31          & 68.88          & 73.60          & 72.51          &\multicolumn{1}{c|}{  72.29 }         & 54.63          & 49.21          & 58.13          & 54.75          & 54.64          & 47.03            & 53.07           \\
FedLMG\_SD     & \textbf{77.34} & \textbf{79.94} & \textbf{75.01} & \textbf{71.87} & \textbf{76.69} & \textbf{74.92} & \multicolumn{1}{c|}{ \textbf{75.96}} & \textbf{61.49} & \textbf{51.47} & \textbf{65.28} & \textbf{60.03} & \textbf{59.57} & \textbf{51.14}   & \textbf{58.16}  \\
FedLMG\_MD     & 74.66          & 75.78          & 71.05          & 69.58          & 74.34          & 72.11          & \multicolumn{1}{c|}{ 72.92  }        & 54.42          & 47.83          & 59.85          & 53.94          & 52.96          & 45.15            & 52.36           \\ \Xhline{1.2pt}
                            
\end{tabular}}
\end{table*}

    \subsection{Model Aggregation}
    \label{3.4}

    Based on the synthetic dataset $\hat{\mathbf{X}}$, we proceed to obtain the aggregated model. We employ the idea of distillation to achieve model aggregation and introduce three strategies to obtain the aggregated model: Fine-tuning, Multi-teacher Distillation, and Specific-teacher Distillation.
    
    \textbf{Fine-tuning.}
    As all samples $\hat{\mathbf{x}}^k_i, i = 1, \dots, N, k = 1, \dots, K$ in the synthetic dataset $\hat{\mathbf{X}}$ have their labels $y^k_i$, this strategy refers to directly training an aggregated model using the cross-entropy loss $\mathcal{L}_{CE}$. The specific loss function used during the aggregation process is as follows:
\begin{align}
    \mathcal{L}_{agg}(\hat{\mathbf{x}}^k_i,y^k_i) =& \mathcal{L}_{CE}(\mathcal{F}_{\boldsymbol{\theta}_g}(\hat{\mathbf{x}}^k_i),y^k_i)
\end{align}
    Although this strategy is relatively simple, since the performance ceiling of centralized training also involves training the aggregated model on the client data, we want to emphasize that this strategy is closest to the centralized training, directly reflecting the quality and diversity difference between the synthetic dataset and the original client dataset.
    
    \textbf{Multi-teacher Distillation.}
The second strategy is multi-teacher distillation. The synthetic dataset $\hat{\mathbf{X}}$ serves as a medium of knowledge distillation, utilizing all client classifiers $\mathcal{F}_{\boldsymbol{\theta}_k}$ as teachers to distill their knowledge into the aggregated model. Specifically, the loss function is as follows:
\begin{align}
    \mathcal{L}_{agg}(\hat{\mathbf{x}}^k_i,y^k_i) & = \mathcal{L}_{CE}(\mathcal{F}_{\boldsymbol{\theta}_g}(\hat{\mathbf{x}}^k_i),y^k_i) \nonumber \\
    &+\lambda KL(\mathcal{F}_{\boldsymbol{\theta}_g}(\hat{\mathbf{x}}^k_i)||\frac{\sum_{h=1}^{K}\mathcal{F}_{\boldsymbol{\theta}_h}(\hat{\mathbf{x}}^k_i)}{K})
\end{align}
where $KL$ means Kullback-Leibler divergence, $\mathcal{F}_{\boldsymbol{\theta}_g}$ and $\mathcal{F}_{\boldsymbol{\theta}_h}$ are the aggregated model and client classifiers, $\lambda$ is the weight of distillation loss. This strategy maximally leverages the knowledge from all client classifiers. However, in cases of substantial variations among clients or under the label distribution skew, the teachers from different clients may provide wrong guidance, impacting the performance of the aggregated model. 

    \textbf{Specific-teacher Distillation.}
    The third strategy is specific-teacher distillation. Given that $\hat{\mathbf{x}}^k_i$ with its client ID $k$, we can use the specific teacher model $\mathcal{F}_{\boldsymbol{\theta}_k}$ to achieve model aggregation. The loss function is as follows: 
\begin{align}
    \mathcal{L}_{agg}(\hat{\mathbf{x}}^k_i,y^k_i) & = \mathcal{L}_{CE}(\mathcal{F}_{\boldsymbol{\theta}_g}(\hat{\mathbf{x}}^k_i),y^k_i) \nonumber \\
    &+ \lambda KL(\mathcal{F}_{\boldsymbol{\theta}_g}(\hat{\mathbf{x}}^k_i)||\mathcal{F}_{\boldsymbol{\theta}_k}(\hat{\mathbf{x}}^k_i))
\end{align}
The meanings of each parameter are the same as mentioned earlier.
When there are significant differences between clients or under the label distribution skew, this strategy ensures accurate guidance and the stable performance of the aggregated model.
 
\section{Experiments}

\subsection{Experimental Settings}
\textbf{Datasets and Implementation Details.}
We conduct experiments on three large-scale real-world image datasets: \textbf{OpenImage}~\cite{kuznetsova2020open}, \textbf{DomainNet}~\cite{peng2019moment} and \textbf{NICO++}~\cite{zhang2022NICO++}. NICO++ can be divided into the common contexts (\textbf{Common NICO++}) and the unique contexts (\textbf{Unique NICO++}). Due to space constraints, we provide comprehensive descriptions and other details in the appendix.
  \begin{figure}[t]
 \centering
 \includegraphics[width=\linewidth]{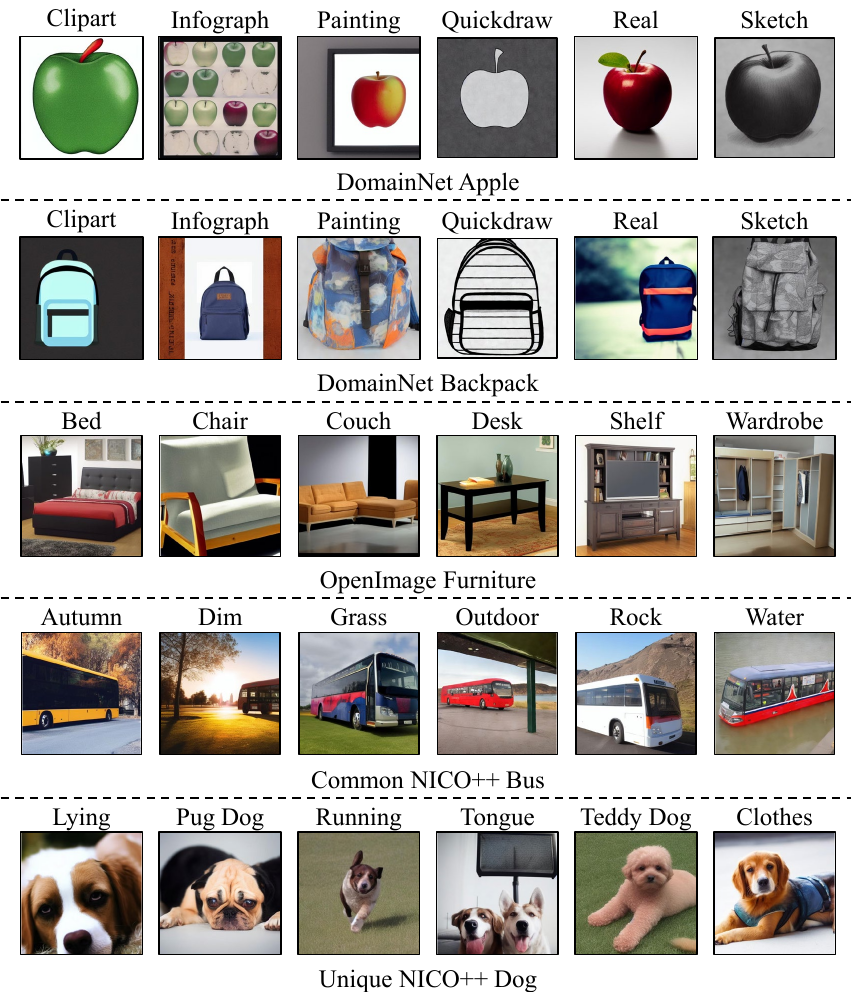}
 \caption{The visualization of generated samples on different datasets. }
 \label{vis_re}
 \end{figure}
 
\textbf{Compared Methods.}
We primarily compare 3 strategies of FedLMG: Fine-Tuning (FedLMG\_FT), Multi-teacher Distillation (FedLMG\_MD), and Specific-teacher Distillation (FedLMG\_SD) against 3 kinds of methods: 1) \textbf{Ceiling.} The performance ceiling of traditional FL methods is centralized training, involving the uploading of all client local data for the training of the aggregated model. 2) Traditional FL methods with multiple rounds of communications: \textbf{FedAvg}~\cite{mcmahan2017communication}, \textbf{FedDF}~\cite{lin2020ensemble}, \textbf{FedProx}~\cite{li2020federated}, \textbf{FedDyn}~\cite{acar2021federated}. All of them have 20 rounds of communications. Following standard experimental settings, each round involves one epoch of training on each client. And we use ImageNet as the additional public data for distillation in FedDF. 3) Diffusion-based OSFL methods: \textbf{FedDISC}~\cite{yang2023exploring}, \textbf{FGL}~\cite{zhang2023federated} and \textbf{Prompts Only}. Although FedDISC is designed for semi-supervised FL scenarios, we remove the pseudo-labeling process of FedDISC and directly utilize the true labels of client images. Another point to notice is the \textbf{Prompts Only}, where the server does not use the client models uploaded from clients at all but only uses the text prompts of category names in the server image generation.
It is important to note that, we also compare our method with \textbf{DENSE}~\cite{zhang2022dense} and \textbf{FedCVAE}~\cite{heinbaugh2022data}. However, due to the reasons highlighted in the Introduction, these methods primarily demonstrate results on smaller datasets like CIFAR-10. Therefore, these methods are not utilized here.

 \begin{table*}[t]
 \caption{The comparison about the communication costs.}
\label{commu}
\center
\resizebox{0.98\linewidth}{!}{
\begin{tabular}{ccc|ccc|ccc|ccc|ccc}
\Xhline{1.2pt}
\multicolumn{15}{c}{Parameters requiring communication (M)}
\\ \hline
\multicolumn{3}{c|}{Ceiling} & \multicolumn{3}{c|}{FedAvg} & \multicolumn{3}{c|}{FedDISC} & \multicolumn{3}{c|}{FGL}    & \multicolumn{3}{c}{FedLMG} \\ \hline  
Upload  & Download & Total  & Upload & Download & Total  & Upload  & Download & Total  & Upload & Download & Total  & Upload  & Download  & Total \\
270.95  & 0        & 270.95 & 233.8  & 222.11   & 455.91 & 4.23    & 427.62   & 431.85 & 0.345   & 469.73   & 470.08 & 11.69   & 0         & \textbf{11.69}
\\
\Xhline{1.2pt}
\end{tabular}}
\end{table*}

\begin{table}[t]
\caption{The comparison of the client computation costs.}
\label{compu}
\center
\resizebox{0.8\linewidth}{!}{
\begin{tabular}{ccccc}
\Xhline{1.2pt}
\multicolumn{5}{c}{Client computation costs (M)}
\\ \hline
& FedAvg               & FedDISC              & FGL                  & FedLMG              \\ \hline
flops (G)            & 72.8                 & 334.73               & 227.34               & \textbf{3.64}                 \\

\Xhline{1.2pt}

\end{tabular}}
\end{table}

\subsection{Main Results}
Firstly, we conduct experiments to assess our method under feature distribution skew. From the results in Table~\ref{fea_skew}, we highlight several observations: 1). Compared to all used methods, FedLMG consistently demonstrates superior performance across all datasets, which effectively demonstrates the performance of our method on large-scale realistic datasets.
2). Compared to Ceiling, in multiple data domains, our method exhibits superior performance, which confirms that the vast knowledge of diffusion models can effectively assist in the training of the aggregated model, resulting in the performance surpassing the traditional performance ceiling of centralized training.
3). Compared to Prompts Only, FedLMG shows promising performance on most clients, which emphasizes the necessity of assistance from the client models. Images generated solely based on text prompts have a distribution that is too broad and cannot comply with the local image distribution of the clients.
4). Compared to other diffusion-based methods, FedLMG demonstrates a performance advantage on most clients. This indicates that our method can extract more precise information about the client distribution from the client models, guiding the DM to generate higher-quality synthetic datasets.
5). The comparison of three model aggregation strategies shows that FedLMG\_SD achieves superior performance on most clients, further confirming the ability of our method to generate data that complies with different distributions, enabling the specific teachers to provide more accurate guidance. Additionally, on DomainNet, the reason FedLMG\_MD performs better on more clients is that the distribution within the same data domain is more complex, and there is more overlap between domains, allowing teachers from other clients to contribute valuable information.

To validate the generating ability of our method, we present visualization results in Figure \ref{vis_re}, illustrating that FedLMG successfully generates images that possess accurate semantic information and exhibit competitive quality with the original client datasets. Due to space limitations, please refer to the appendix.

\begin{table}[t]
\label{ablation}
\caption{The influence of different losses.}
\center
\resizebox{1.0\linewidth}{!}{
\begin{tabular}{ccccccccc}
\toprule
\begin{tabular}{c}{{}}BN Loss \end{tabular}  & \begin{tabular}{c}{{}}CLS Loss\end{tabular} 
 &  clipart        & infograph      & painting    &quickdraw   & real           & sketch & average \\ \midrule
 &  & 31.8  & 11.61 & 31.14 & 4.13 & 61.53 & 31.44 & 28.60 \\ 
$\checkmark$ &  & 40.55 & 15.89 & 36.84 & 7.64 & 58.87 & 36.05 & 32.64 \\ 
 & $\checkmark$ & 38.92 & 14.82 & 33.57 & 4.61 & 58.69 & 36.07 & 31.11 \\ 
$\checkmark$ & $\checkmark$ & \textbf{44.25} & \textbf{17.51} & \textbf{38.74} & \textbf{9.43} & \textbf{57.31} & \textbf{38.44} & \textbf{34.28}  \\ \bottomrule
\end{tabular}}
\end{table}

\subsection{Ablation Experiments}
We conduct extensive ablation experiments to validate the effects of hyperparameters and components of our method. Due to space limitations, please refer to the supplementary material for most of the results.

\begin{figure}[t]
 \centering
 \includegraphics[width=\linewidth]{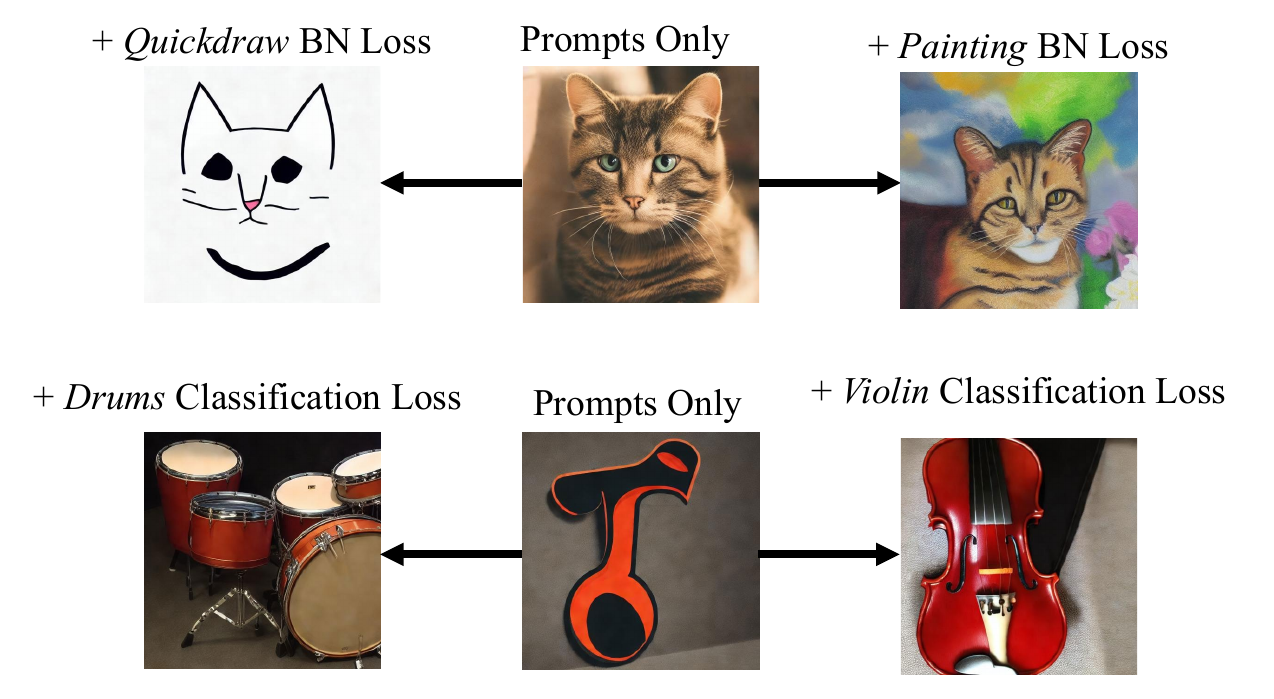}
 \caption{The visualization about the effects of different loss functions. }
 \label{vis_abla}
 \end{figure}

\textbf{The ablation experiments about the loss functions.}
Table~\ref{ablation} and Figure~\ref{vis_abla} demonstrate the roles of the BN loss and classification loss. As shown in Table~\ref{ablation}, the inclusion of BN loss and classification loss introduces category and contextual information, leading to the generation of synthetic datasets that better comply with client distributions, thereby training an aggregated model with improved performance. Figure~\ref{vis_abla} illustrates that classification loss provides more precise category information during the generation process, while BN loss introduces more detailed contextual information, preserving the stylistic alignment between the generated dataset and the client data.

\subsection{Discussions and Limitations}

\textbf{Communication Costs.}
We thoroughly discuss the communication cost of the proposed method. Since the communication costs of FedAvg, FedDF, FedProx, and FedDyn are essentially the same, their results are not repeated. Prompts Only does not involve any communication between the client and the server. The number of iterations and the used model structures follow the default experimental settings. The comparison results of the upload and download communication costs between FedLMG and other methods are shown in Table~\ref{commu}. From the results, it is evident that because there is no foundation model used on the clients, FedLMG does not involve any download communication cost, resulting in the lowest communication.

\textbf{Computation Costs.}
The computation costs include the computation costs on the client and the server. Regarding the server computation costs, on one hand, as same as other diffusion-based OSFL methods, FedLMG involves generating synthetic datasets and training the aggregated model on the server, leading to similar server computation costs. On the other hand, in FL, although generating data requires more computation on the server, the server, as the center of the FL, typically has sufficient computational power. Therefore, lower client computation costs are relatively more advantageous for practicality~\cite{kairouz2021advances}. Regarding client computation costs, we conduct thorough quantitative experiments in Table~\ref{compu}. Since FedAvg, FedDF, FedProx, and FedDyn have similar client computation costs, they are not compared separately. Ceiling and Prompts Only do not involve any client computation, so they are not included in the comparison. The number of iterations and the model structures used follow the default experimental settings. The quantified results demonstrate that FedLMG has a significant advantage in client computation costs due to not involving any foundation model on the clients and only requiring training of client models in a single round of iteration, demonstrating its practicality.

\textbf{Privacy Issues.}
 \begin{figure}[t]
 \centering
 \includegraphics[width=\linewidth]{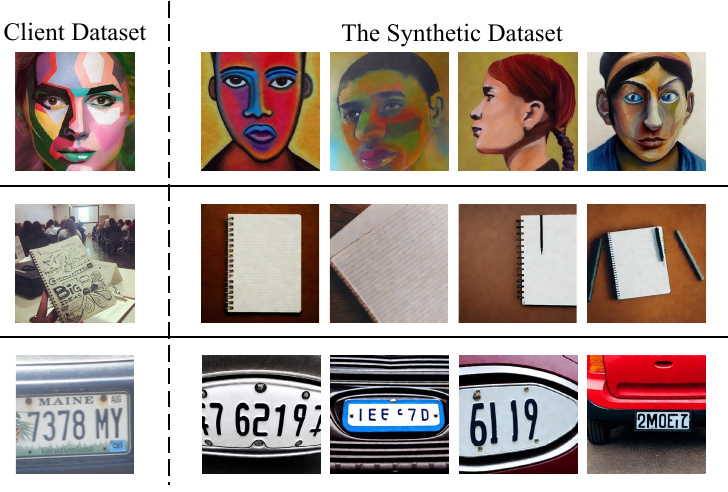}
 \caption{The visualization of privacy-sensitive information-related categories. }
 \label{vis_priv}
 \end{figure}
Transmitting client models is the most common practice in FL. Since our method only requires uploading the client model once, it offers a significant advantage in privacy protection compared to other traditional FL methods. Compared to other OSFL methods, where either the trained generative model~\cite{zhang2022dense,heinbaugh2022data} or direct descriptions of client images are uploaded~\cite{yang2023exploring,zhang2023federated}, extracting user privacy information from client models is more challenging.
 To further demonstrate FedLMG's performance in privacy protection, we conduct sufficient quantitation and visualization experiments. We select some categories from OpenImage that may contain privacy-sensitive information, such as human faces, vehicle registration plates, and notebooks. We train client models on these categories and generate synthetic datasets. The visualization results are shown in Figure~\ref{vis_priv}. It can be observed that the synthetic datasets only share similar styles and accurate semantics with the original client datasets. It is almost impossible to extract specific privacy-sensitive information from the client models, which aim to learn the classification boundary. Due to space limitations, please refer to the appendix for detailed experiments about privacy issues.

\section{Conclusion}

In this paper, we propose FedLMG. Compared to existing OSFL methods, we eliminate the need for auxiliary datasets and generator training, making it effortlessly applicable in real-world scenarios. Comprehensive experiments on three large-scale datasets demonstrate that the proposed FedLMG outperforms all compared methods and even surpasses the performance ceiling of centralized training in some cases, underscoring the potential of applying DMs in OSFL.

\section*{Impact Statement}
Given that our method utilizes a pre-trained diffusion model, there is a possibility of generating sensitive or private information. However, the Stable Diffusion model we rely on has been equipped with a robust safety-checking mechanism designed to minimize such risks. We also conduct sufficient experiments to demonstrate the performance of our method in privacy protection. As a result, we feel that our method does not raise additional significant concerns or potential broader impacts that warrant specific attention or further discussion here.

\bibliography{example_paper}
\bibliographystyle{icml2025}

\end{document}